\documentclass[conference]{IEEEtran}
\IEEEoverridecommandlockouts

\usepackage{cite}
\usepackage{amsmath,amssymb,amsfonts}
\usepackage{algorithmic}
\usepackage{graphicx}
\usepackage{textcomp}
\usepackage{xcolor}
\usepackage{hyperref}
\usepackage{array}
\usepackage{tabularx}
\usepackage{dblfloatfix}
\usepackage{fancyhdr}

\hypersetup{%
	pdfborder = {0 0 0}
}
\def\BibTeX{{\rm B\kern-.05em{\sc i\kern-.025em b}\kern-.08em
    T\kern-.1667em\lower.7ex\hbox{E}\kern-.125emX}}
\begin{document}
\makeatletter
\def\ps@IEEEtitlepagestyle{%
	\def\@oddfoot{\mycopyrightnotice}%
	\def\@evenfoot{}%
}
\def\mycopyrightnotice{
	{\footnotesize 979-8-3503-8848-0/24/\$31.00~\copyright2024 IEEE \hfill}
	\gdef\mycopyrightnotice{}
}
\title{An Evolutionary Large Language Model for Hallucination Mitigation
}

	\author{\IEEEauthorblockN{Abdennour Boulesnane$^*$ \href{https://orcid.org/0000-0002-2272-4953}{\includegraphics[scale=0.02]{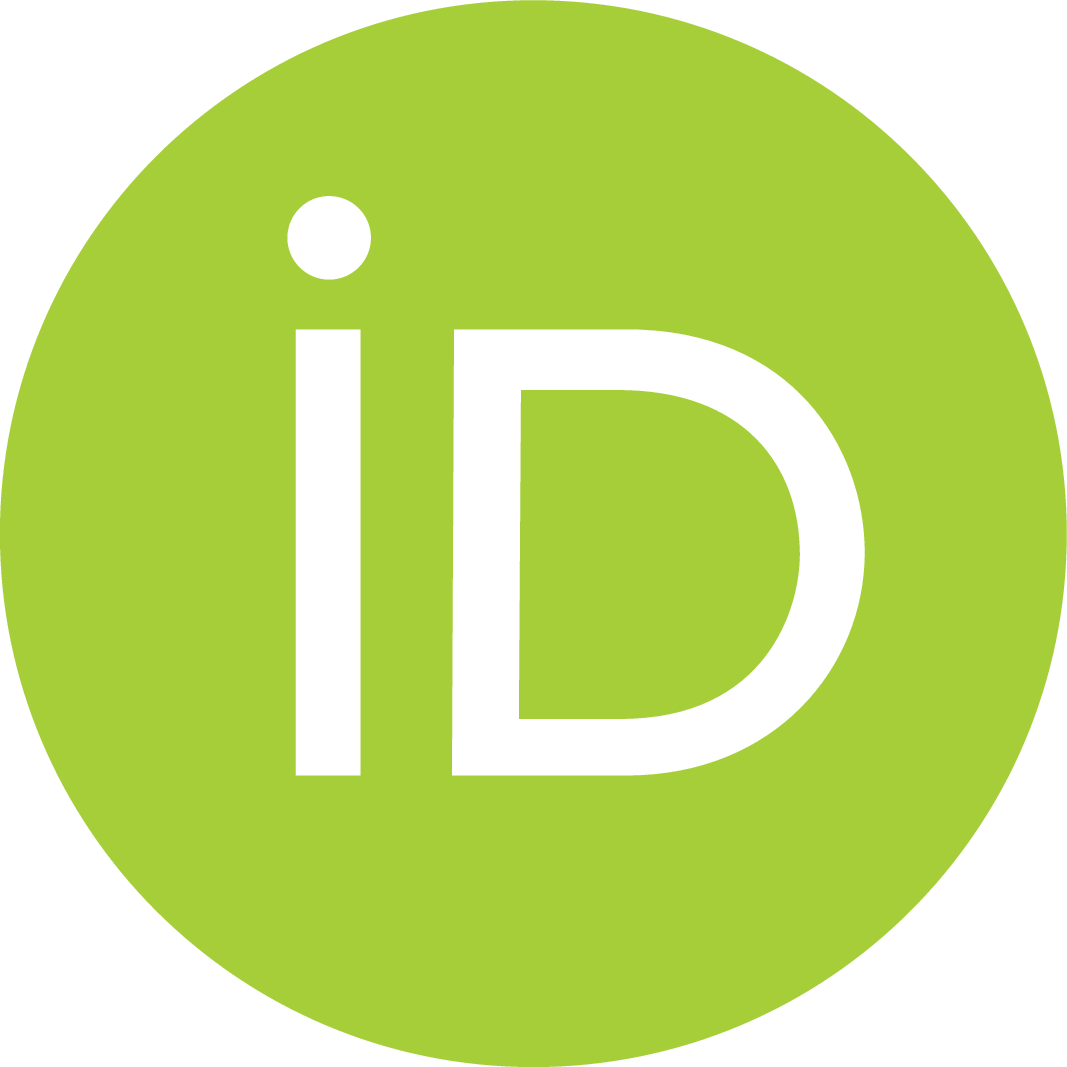}}}
	\IEEEauthorblockA{BIOSTIM Laboratory \\
		Faculty of Medicine \\
		Salah Boubnider University\\
		Constantine, Algeria \\
		aboulesnane@univ-constantine3.dz}
	\and
\IEEEauthorblockN{Abdelhakim Souilah$^*$ \href{https://orcid.org/0009-0002-5713-8863}{\includegraphics[scale=0.02]{figs/orcid.eps}}}
	\IEEEauthorblockA{Department of IFA\\
		Faculty of NTIC \\
		Abdelhamid Mehri University\\
		Constantine, Algeria \\
		abdalhekim.souilah@univ-constantine2.dz}
}

\maketitle
\def\thefootnote{*}\footnotetext{The authors contributed equally to this work.}
\renewcommand{\thefootnote}{\arabic{footnote}}
\lfoot{\mycopyrightnotice}
\begin{abstract}
The emergence of LLMs, like ChatGPT and Gemini, has marked the modern era of artificial intelligence applications characterized by high-impact applications generating text, images, and videos. However, these models usually ensue with one critical challenge called hallucination: confident presentation of inaccurate or fabricated information. This problem attracts serious concern when these models are applied to specialized domains, including healthcare and law, where the accuracy and preciseness of information are absolute conditions. In this paper, we propose EvoLLMs, an innovative framework inspired by Evolutionary Computation, which automates the generation of high-quality Question-answering (QA) datasets while minimizing hallucinations. EvoLLMs employs genetic algorithms, mimicking evolutionary processes like selection, variation, and mutation, to guide LLMs in generating accurate, contextually relevant question-answer pairs. Comparative analysis shows that EvoLLMs consistently outperforms human-generated datasets in key metrics such as Depth, Relevance, and Coverage, while nearly matching human performance in mitigating hallucinations. These results highlight EvoLLMs as a robust and efficient solution for QA dataset generation, significantly reducing the time and resources required for manual curation.
\end{abstract}


	\section{Introduction}
With its introduction in November 2022, ChatGPT set a new milestone for Artificial Intelligence, catapulting transformer-based generative AI into the limelight \cite{Roumeliotis2023}. This transformative technology can create text, images, and even videos. Considerable interest and investment by tech giants, such as Microsoft, Google, and Nvidia, have made generative AI one of the crucial drivers of the future of AI \cite{Roumeliotis2023}. Generative AI is a source of highly realistic and captivating content that can spark unprecedented interest and innovation in this area.

Despite their remarkable capabilities, Large Language Models (LLMs) face a notable challenge known as "hallucination," where the models confidently provide wrong or fabricated information \cite{HuangHuang2023}. This problem becomes particularly concerning in specialized fields, as evidenced by numerous inaccurate responses when ChatGPT is used for health-related inquiries. The prevalence of such errors reveals the urgent need for domain LLMs that are credible and accurate in providing information, especially in sensitive domains involving healthcare, where misinformation can lead to serious consequences.

Moreover, the development of AI applications largely relies on the availability of high-quality datasets, especially in domains requiring specialized knowledge, such as question-answer datasets \cite{NEURIPS2023_9cb2a749}. Question-answering systems, which also rely on LLMs for accurate and relevant responses, find a growing application in healthcare, legal, and other support services \cite{Ojokoh2019}. However, their full effectiveness is still hindered by the lack of adequate volumes of high-quality domain-specific data. For this reason, when LLMs operate on incomplete or scarce information, they often produce what's referred to as hallucination, a fabrication or incorrect information made to sound credible.\\
The problem is compounded by the fact that most traditional methods of collecting data involve manual curation, done by human experts, usually very slow and expensive. Constructing such datasets requires formulating meaningful questions, sourcing relevant information, and formulating correct answers, all highly time-consuming and requiring copious amounts of expertise. Additionally, the financial cost in these processes is very high, more so when the resources available for the projects are minimal. Besides, human involvement could contribute a degree of bias. Such biases could come with the selection of questions and answers that reflect the creators' subjective views or even the cultural background in which the datasets are supposed to serve.

To tackle these challenges, this paper presents novel methods that harness the capabilities of LLMs to generate QA datasets while minimizing the issue of hallucinations efficiently. The proposed framework draws inspiration from Evolutionary Computation (EC), particularly genetic algorithms, to automate the contextual generation of QA data using LLMs, such as Google's Gemini Pro \cite{GeminiTeam}. By mimicking evolutionary processes like selection, crossover/variation, and mutation, the framework leverages the generative power of LLMs to enhance QA system performance while reducing the risk of hallucinations.\\
A comparative analysis was done, and the QA pairs generated by our EvoLLMs consistently outperformed those curated by human experts in all metrics. This further corroborates the robustness of our framework in reducing hallucinations while maintaining the accuracy and relevance of the generated content.

The paper is structured as follows: Section \ref{sect2} introduces LLMs and the issue of hallucinations. Section \ref{sect3} reviews related work on mitigating hallucinations. Section \ref{sect4} explores the synergy between LLMs and evolutionary computation. Section \ref{sect5} presents the proposed framework and methodology. Section \ref{sect6} analyzes the results and discusses their implications. Finally, Section \ref{sect7} offers conclusions and suggests directions for future research.
\section{LLMs and Hallucinations} \label{sect2}
Large Language Models (LLMs) are deep artificial intelligence models that generate and understand human language; they are trained on large amounts of text data in bulk \cite{Naveed2023}. The Transformer architecture is utilized in these models, which lets them deal with sequential data much more effectively \cite{Roumeliotis2023}. The self-attention mechanism of Transformers allows LLMs to learn such complicated relations within text that the models can produce coherent and contextually relevant responses \cite{Chang2024}.
LLMs, such as GPT-3, GPT-4, and BERT, represent a significant leap from traditional language models by incorporating massive parameters and training on large-scale datasets. This expansion has led to improvements in performance and versatility across various natural language processing tasks. For example, GPT-3 and GPT-4 are known for generating human-like text and performing complex language tasks, while BERT excels in understanding the context within text \cite{Raiaan2024}.
\begin{figure}[htbp]
	\centering 
	\includegraphics[width=\columnwidth]{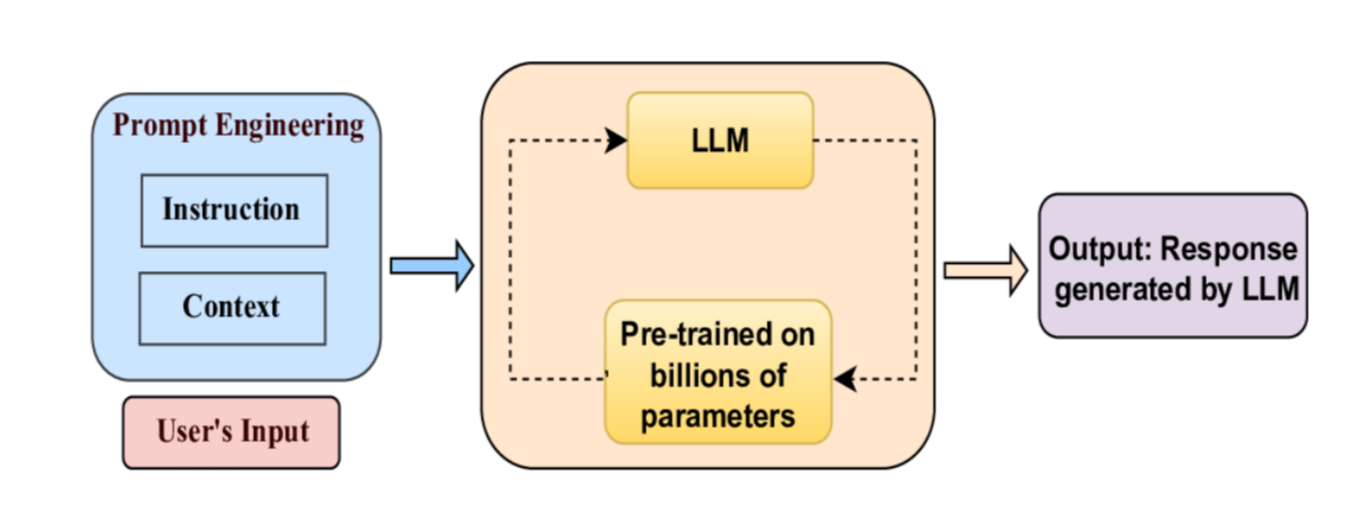} 
	\caption{Prompt engineering guides LLM response generation \cite{Sahoo2024}.} 
	\label{fig1} 
\end{figure}

Furthermore, the effectiveness of LLMs is highly dependent on the prompts. Prompt engineering, as highlighted by Sahoo et al. \cite{Sahoo2024}, is a crucial technique for unlocking the full potential of LLMs by crafting carefully designed input queries that guide the model toward specific tasks (see Figure \ref{fig1}). This approach bridges user intent and LLM output, enabling the model to perform a wide range of actions without extensive retraining.\\ Techniques such as zero-shot and few-shot prompting address new tasks, chain-of-thought (CoT) prompting aids in complex reasoning, and self-consistency and logical chain-of-thought (LogiCoT) prompting enhance verification and structured reasoning. Additionally, Reinforcement Learning from Human Feedback (RLHF) is employed to fine-tune these models, allowing them to learn from human corrections and improve over time. This rapidly evolving field of prompt engineering promises a future where LLMs seamlessly integrate into our lives, tackling complex tasks and providing tailored solutions \cite{Marvin2024}.\\

Despite their advanced capabilities, LLMs encounter several significant challenges \cite{Chen2024}. The process of generating textual datasets can be slow and cumbersome due to the models' immense size and complexity, further complicated by their limited capacity to handle extended inputs effectively. Their reliance on prompts introduces a notable vulnerability; even minor modifications in input can lead to drastically different outputs, resulting in inconsistent and unpredictable responses.

Moreover, LLMs are prone to "hallucinations", a phenomenon where the model produces information that appears plausible but is either inaccurate or entirely fabricated \cite{HuangHuang2023}. This issue is particularly concerning in critical fields such as healthcare, where misinformation can have dire consequences. For example, an LLM might erroneously recommend a potentially harmful treatment or dissuade from a well-established and effective therapy, thereby endangering patient safety. Addressing these challenges, ensuring that LLM outputs align with human values, preventing the spread of misinformation, and continually updating knowledge to avoid reliance on outdated data, remains a pressing and ongoing concern.

\section{Related Work on Mitigating Hallucinations} \label{sect3}
Minimizing hallucinations in LLMs is crucial, as it significantly affects the trustworthiness and reliability of the generated content. This section will review the most recent efforts to mitigate hallucinations in LLMs. Given the growing complexity and widespread use of LLMs, addressing hallucination instances where the model generates factually incorrect or misleading information has become a critical area of research. This overview will focus on the latest advancements, techniques, and methodologies to reduce these inaccuracies and improve model reliability. 

One effective method for tackling hallucinations in LLMs is incorporating external knowledge using techniques like Retrieval Augmented Generation (RAG) \cite{Patrice08189, Kirchenbauer2024}. RAG has proven to be a valuable approach in prompt engineering to mitigate hallucinations. It works by accessing external knowledge bases to provide the LLM with current and relevant information, which enhances its ability to produce evidence-based responses \cite{Tonmoy2024}.

In a different work by Guan et al. \cite{Guan2024}, a retrofitting framework is introduced that leverages knowledge graphs (KGs) to reduce factual hallucinations in LLMs. Rather than directly retrieving factual data from KGs using original queries, the proposed approach refines the initial outputs of LLMs by aligning them with verified information stored in KGs. The core aspect of this methodology involves identifying portions of the draft responses that require validation, retrieving relevant knowledge from the KGs, and utilizing this information to verify and adjust the generated responses, thus enhancing the factual accuracy of the model's reasoning process.

In the study by McDonald et al., \cite{McDonald2024}, knowledge distillation is explored to address hallucination in LLMs. This approach focuses on transferring the knowledge and behaviors of a larger, more complex model to a smaller, more efficient one to maintain the original model's performance while improving computational efficiency. The results demonstrated that the distilled model outperformed the baseline in accuracy and contextual relevance, highlighting the effectiveness of knowledge distillation in refining model outputs and reducing hallucination.

Suzuoki et al. \cite{Suzuoki2024} introduced a majority voting mechanism within a Mixture of Experts (MoE) framework to substantially reduce hallucinations in LLMs. The effectiveness of this method was evaluated through controlled experiments, comparing the hallucination rates of the baseline model with those of the model incorporating majority voting. The findings revealed a notable decrease in hallucinations when the majority voting mechanism was employed, underscoring its potential to enhance the reliability of language model outputs.

Fairburn et al. \cite{Fairburn2024} explored the integration of Graph Neural Networks (GNN) with LLMs to reduce hallucinations and enhance contextual comprehension across various tasks. The approach modifies the Llama 2 architecture by incorporating a GNN component, which processes relational data produced by the LLM. This GNN element refines the outputs using relational structures, improving factual accuracy and coherence. Experimental results showed significant enhancements in the hybrid GNN-LLM model, reflecting increased reliability and precision.

Ainsworth et al. \cite{Ainsworth2024} focus on optimizing attention maps to minimize contextual hallucinations in LLMs. Utilizing the Llama model as a foundation, they developed a technique that analyzes attention maps to detect hallucination-related patterns and adjusts attention weights dynamically during the inference process. The results from their experiments demonstrated that this approach improves the model's ability to align outputs with factual information while maintaining contextual coherence.

Authors in \cite{Desrochers2024} propose a new technique based on Contextual Position Encoding (CPE), which dynamically encodes positional information relative to the context of each token, greatly improving the model's capability to produce coherent and accurate text. Integrating CPE into the hidden layers of Mistral Large, the researchers conducted a comparative analysis with baseline models like GPT-3 and BERT. The results demonstrated that CPE outperformed these models, showcasing its effectiveness in enhancing text generation accuracy.

The work in \cite{Huang2024} introduces a method to enhance the GPT-Neo autoregressive transformer model by integrating Low-Rank Adaptation (LoRA) to address hallucinations and boost factual accuracy. LoRA breaks down the model's weight matrices into lower-rank approximations, reducing computational and memory requirements while maintaining the model's original hierarchical structure. This modification, applied to the attention and feed-forward layers, facilitates more efficient parameter updates and better contextual accuracy. The adapted model showed notable improvements in BLEU and ROUGE-L scores, reflecting enhanced text quality compared to the baseline models.

The study in \cite{Lin20505} explores a multi-agent debate approach by deploying several large LLMs as agents to engage in a series of debate rounds. In this method, when Agent A provides an answer, Agent B evaluates and critiques it. Agent A then reassesses its response based on Agent B's feedback. After multiple debate rounds, a designated Agent Judge delivers a final verdict if a consensus is not achieved. This debate process helps identify hallucinations and provides insights into their causes and specifics.

Hu et al. \cite{Hu11267} present Faithful Finetuning (F2) for Question Answering (QA), a method designed to enhance response accuracy using specialized loss functions during finetuning. They split the QA objective into internal fact retrieval and fact-grounded QA, which helps the model use its knowledge more effectively. Their approach includes targeted finetuning of critical areas identified through entity-based and attention-based heuristics and specifically addresses layers prone to hallucinations. The results show that F2 significantly improves performance over other models.

For a comprehensive understanding of hallucination mitigation in LLMs, readers are encouraged to refer to recent survey papers, such as those in \cite{Tonmoy2024, Luo2024} for a more in-depth exploration and advancements in the field.
\section{LLMs and Evolutionary Computation}\label{sect4}
Evolutionary computation (EC) has emerged as a vital field within computer science, gaining significant traction due to its advantages over traditional deterministic methods. EC draws inspiration from biological evolution and adaptation, applying these natural principles to computational systems to develop effective solutions for complex real-world problems \cite{nayyar2018evolutionary}.
 \begin{figure}[b!]
	\centering 
	\includegraphics[width=\columnwidth]{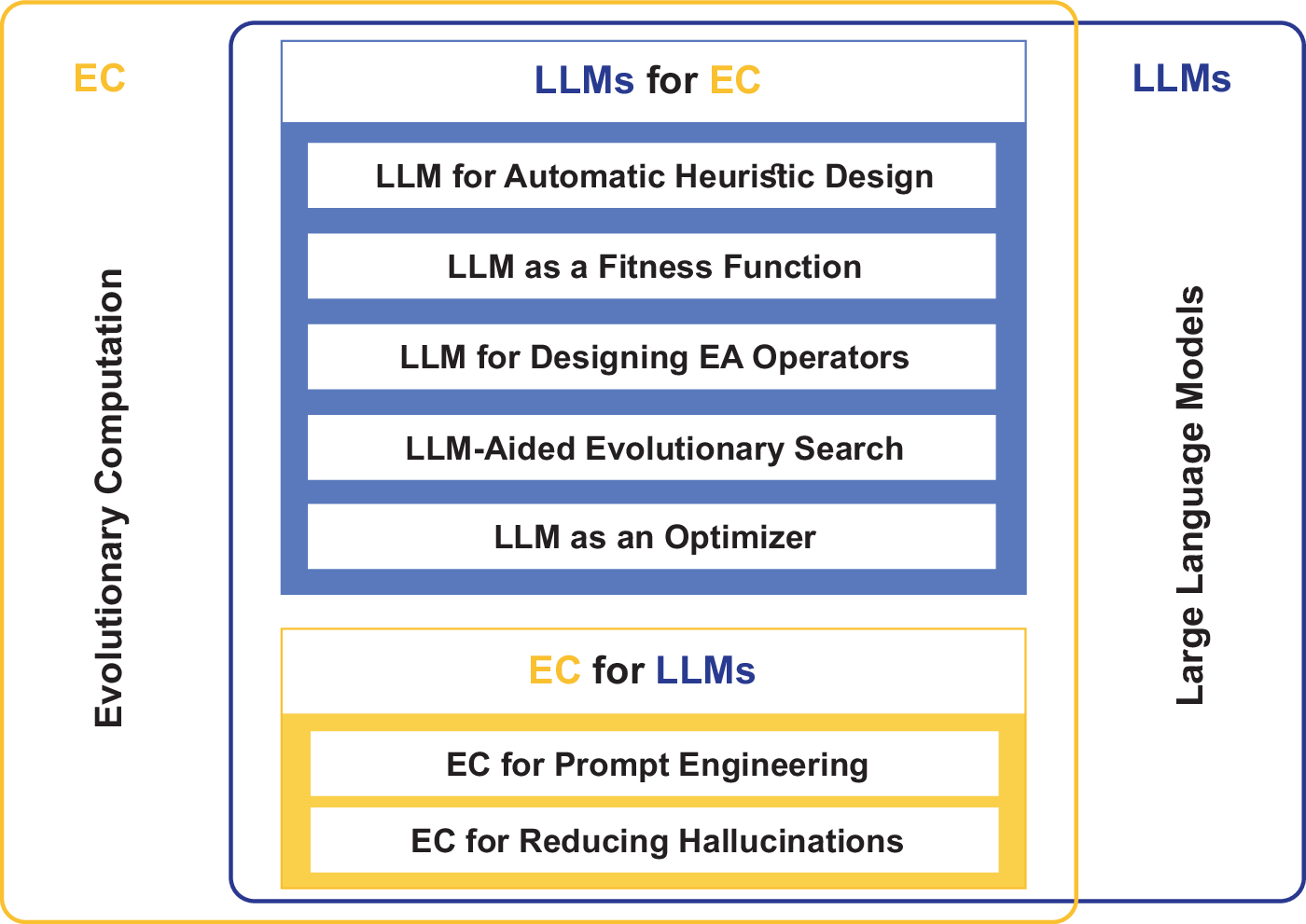} 
	\caption{Synergy of Evolutionary Computation and LLMs.} 
	\label{fig2} 
\end{figure}
\begin{table*}[!bp]
	\centering
	\caption{Using Genetic Algorithm Concepts as an Inspiration and Analogy for Data Generation in EvoLLMs.}
	
	\setlength{\extrarowheight}{2.5pt}
	\begin{tabularx}{0.95\textwidth}{|l|X|X|}
		
		\hline
		\textbf{Genetic Algorithm Concepts} & \textbf{Implementation Analogy} & \textbf{Explanation} \\ \hline
		\textbf{Initial Population} & Initial QA Generation & Model 1 (Gemini Pro 1.5) generates 10 unique QA pairs from the PDF document based on the context and structured prompts. \\ \hline
		\textbf{Selection} & QA Pair Evaluation & Model 3 evaluates each QA pair based on 15 metrics to select the highest-scoring pairs, similar to selecting the fittest individuals. \\ \hline
		\textbf{Crossover} & Variation Generation & Model 2 creates 10 variations of each QA pair, akin to crossover, combining and altering elements to generate diverse offspring. \\ \hline
		\textbf{Mutation} & Iterative Refinement & Introducing new variations of QA pairs ensures diversity and prevents stagnation, similar to mutations introducing new genetic material. \\ \hline
		\textbf{Fitness Function} & Quality Metrics Evaluation & Model 3's evaluation process uses various metrics (relevance, depth, accuracy) as a fitness function to assess and score QA pairs. \\ \hline
		\textbf{Elitism} & Threshold Passing & QA pairs that score 8/10 or higher are directly included in the final dataset, preserving high-quality individuals in the population. \\ \hline
		\textbf{Feedback Loop} & Iterative Generation & If the highest-scoring QA pair falls below the threshold, it is fed back into Model 2 to generate new variations, similar to the iterative nature of genetic algorithms. \\ \hline
		\textbf{End Criteria} & Threshold Achievement & The iterative process continues until QA pairs meet the quality threshold, akin to terminating a genetic algorithm when an optimal solution is found. \\ \hline
	\end{tabularx}
	\label{table:comparative}
\end{table*}
By incorporating mechanisms similar to those seen in evolution, such as selection, mutation, and reproduction, EC algorithms aim to create adaptive systems that evolve toward optimal solutions. These evolutionary algorithms (EAs), including popular forms like genetic algorithms, operate through a population of potential solutions refined through fitness criteria and genetic operators \cite{KanakaVardhini2016}.
 
As shown in Figure \ref{fig2}, the synergy between LLMs and EC represents a powerful fusion of natural language processing and evolutionary algorithms \cite{Wu240110034}. LLMs could play several critical roles in enhancing EC. One key application is LLM-driven automatic heuristic design \cite{Liu02051}, where LLMs generate and refine heuristics by learning from large datasets, uncovering patterns in problem-solving, and creating tailored solutions to specific optimization challenges. Additionally, LLMs can function as dynamic fitness functions \cite{Hao10675}, evaluating potential solutions based on learned criteria. This allows for more adaptive and flexible fitness evaluations, capturing complex relationships that traditional methods may miss. Furthermore, LLMs assist in designing evolutionary algorithm (EA) operators \cite{Huang08987} by creating advanced mutation, crossover, and selection mechanisms suited to particular domains, speeding up the convergence toward optimal solutions.\\
Moreover, in LLM-aided evolutionary searches \cite{Wang2024}, LLMs help guide exploration by identifying promising areas within the search space, enhancing the exploration-exploitation balance, and discovering high-quality solutions more efficiently. Finally, LLMs serve as optimizers \cite{Lange2024} by fine-tuning parameters, predicting optimal configurations, and suggesting new strategies to improve overall performance, making them invaluable tools for optimizing complex environments within EC.

On the other hand, EC can also be leveraged to optimize prompt engineering for LLMs \cite{Saletta2024}. By utilizing evolutionary algorithms to refine and improve prompts iteratively, EC can explore various variations and configurations to identify the most effective input formulations for eliciting desired responses from LLMs.\\
In the context of mitigation hallucinations, EC can be applied to address the issue of hallucinations in LLMs, where the model generates incorrect or fabricated information. EC can optimize prompt structures, response evaluation criteria, or fine-tune model parameters through iterative processes. \\
One of the few studies addressing hallucinations, the noteworthy study in \cite{Kulkarni240900085} tackles the hallucination issue by applying a genetic generative approach that uses generative language models as genetic operators with custom prompts. These operators include random document rewriting, query-specific mutations, and a crossover that merges rewritten documents into a single response. The algorithm takes user queries and text documents as input, with documents serving as genetic representations. A fitness function that ensures relevance and grounding in the source texts selects the best documents for each iteration. Experiments show that this method effectively reduces hallucinations while maintaining relevance to the query.
\section{Framework and Methodology}\label{sect5}
Our paper addresses the critical challenge of high-quality, domain-specific data scarcity in the era of generative AI, particularly in the context of question-answering (QA) systems. QA systems, which rely on LLMs to provide accurate and relevant responses, are increasingly used across various domains, such as healthcare, law, and customer support. However, the scarcity of high-quality, domain-specific data poses a significant risk to the performance of these systems. When faced with incomplete or insufficient data, LLMs can produce hallucinations and fabricated or inaccurate information that appears plausible but lacks factual grounding. These hallucinations undermine the reliability of QA systems, where accuracy is paramount.

\begin{figure}[h]
	\centering 
	\includegraphics[width=\columnwidth]{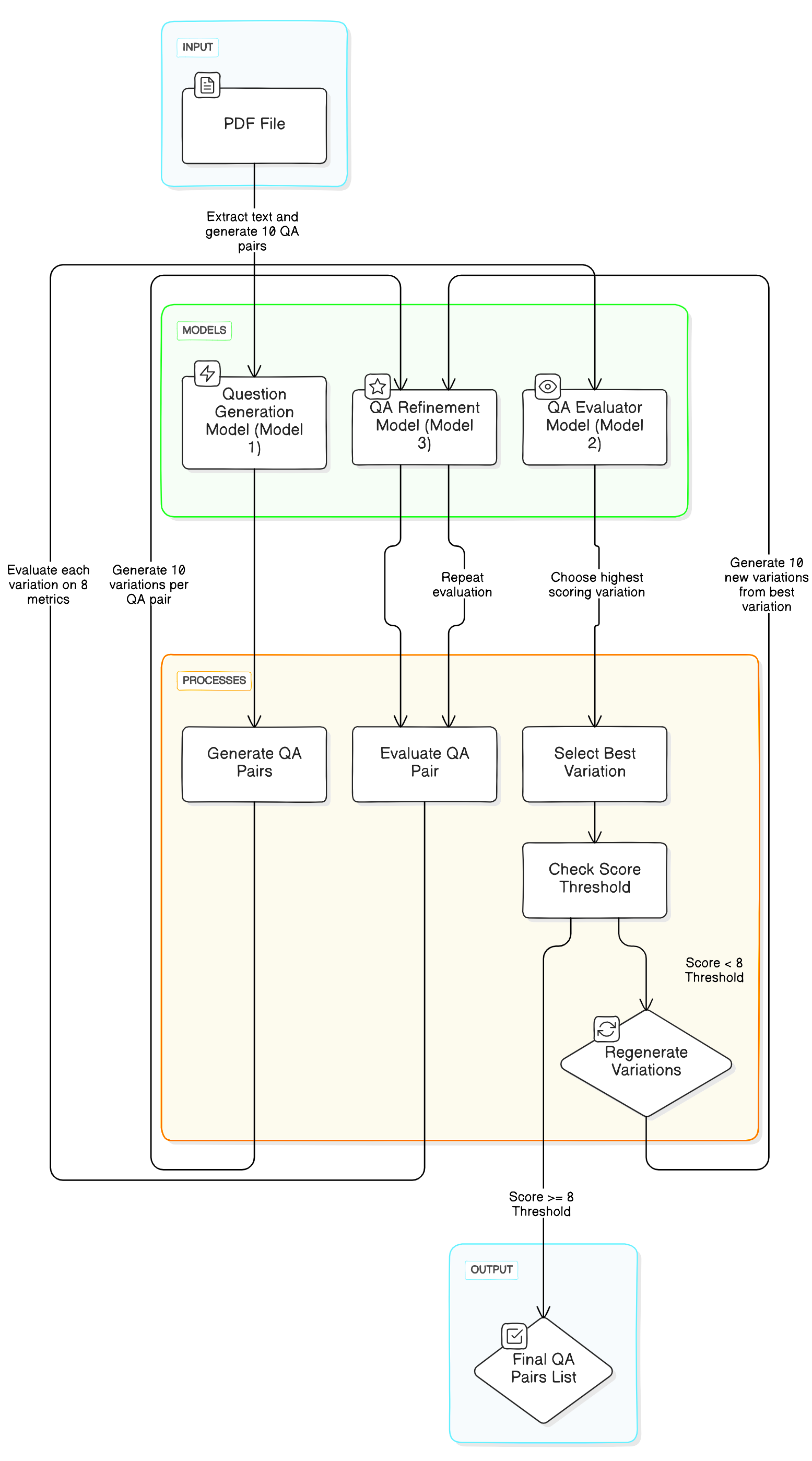} 
	\caption{The Proposed EvoLLMs Framework for Optimizing Question-Answering Systems.} 
	\label{fig3} 
\end{figure}

\begin{figure*}[htp]
	\centering 
	\includegraphics[width=\textwidth]{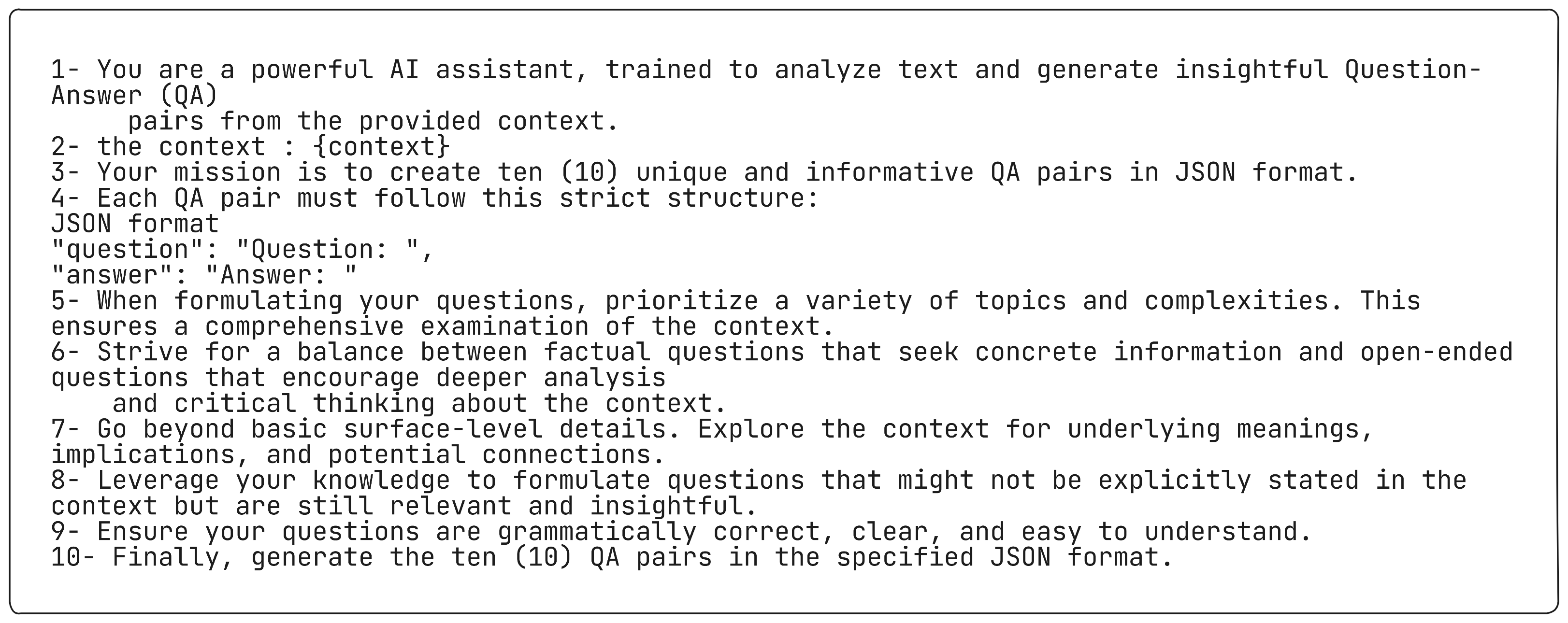} 
	\caption{Prompt 1 for Question-Answer Generation Task.} 
	\label{fig4} 
\end{figure*}

Inspired by EC and genetic algorithms as shown in Table \ref{table:comparative}, we have developed a system for automated, contextualized data generation \cite{Kirchenbauer2024} using LLMs, specifically Google's Gemini Pro \cite{GeminiTeam}. Recognizing that even the most advanced LLMs can produce hallucinations, our EvoLLMs framework incorporates a rigorous iterative refinement process to ensure data accuracy and relevance.
The proposed approach (see Figure \ref{fig3}) centers around mimicking the evolutionary processes of selection, crossover/variation, and mutation, allowing the leverage of the generative power of LLMs while improving the performance of QA systems and mitigating the inherent risk of hallucination.\\
As depicted in Figure \ref{fig3}, the proposed iterative process begins with contextual QA generation, wherein the entire source PDF document is provided as context to a Gemini instance (Model 1) alongside a Chain-of-Thought (CoT) Prompt 1, (see Figure \ref{fig4}). The CoT prompting technique \cite{Wei220111903} encourages the model to generate reasoned, step-by-step responses, thereby reducing the likelihood of hallucinations and promoting factually grounded outputs. Model 1 generates an initial set of 10 unique question-answer pairs, drawing information from different sections of the PDF. This ensures a diverse initial population of candidate QA pairs.

Model 2 (another Gemini instance) generates variations, acting as a source of variation analogous to genetic crossover and mutation, so our second Model receives the QA pairs generated by Model 1 along with Prompt 2. This prompt directs Model 2 to create ten variations of each input QA pair. These variations retain the core semantic information of the parent QA pair while exploring different phrasing, sentence structures, and perspectives. This diversification is crucial for exploring the solution space and avoiding premature convergence on suboptimal QA pairs.

\begin{figure*}[htp]
	\centering 
	\includegraphics[width=\textwidth]{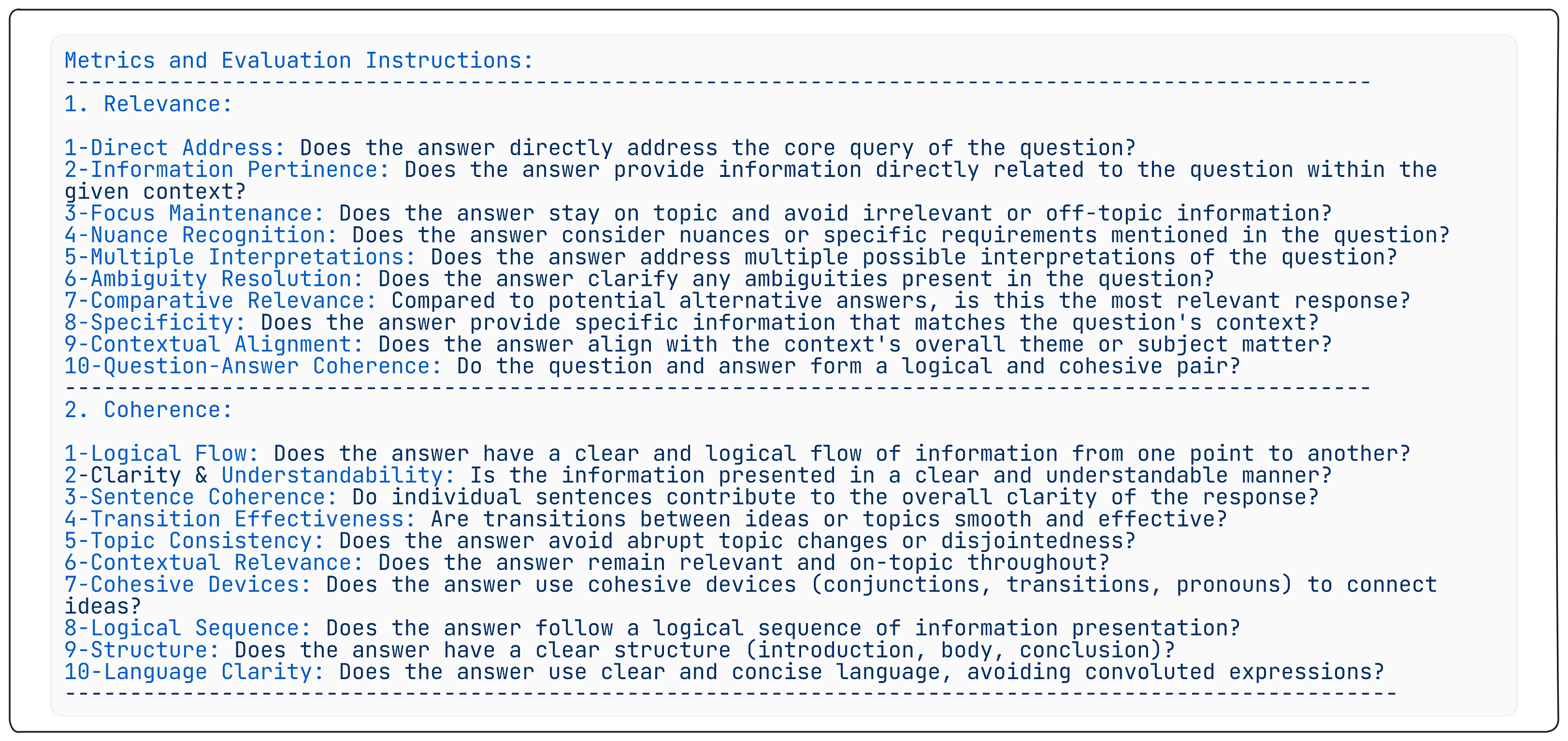} 
	\caption{Evaluation Prompt: Metrics and Evaluation Instructions.} 
	\label{fig5} 
\end{figure*}

The third Model evaluates and selects QAs, mimicking the pressure of natural selection. Model 3, also a Gemini instance, evaluates the generated QA pairs using Prompt 3 (the evaluation prompt, see Figure \ref{fig5}). This prompt leverages the original PDF document as context and provides detailed instructions for assessing each QA pair against 15 distinct quality metrics. These metrics encompass aspects such as relevance to the source document, depth of understanding, factual accuracy, conciseness, clarity, and absence of hallucinations. Each metric contributes to an overall quality score for each QA pair. This evaluation process serves as a fitness function, guiding the evolutionary process towards higher-quality QA pairs.

The QA pair with the highest overall score from Model 3 undergoes a threshold test. A score of 8/10 or higher signifies that the pair meets the predefined quality criteria and is added to the final dataset. If the highest-scoring pair fails to reach the threshold, it is fed back into Model 2 for further variation generation, restarting the cycle. This iterative refinement process continues until a satisfactory QA pair is generated for each initial seed from Model 1. This feedback loop, analogous to generational evolution, drives the system towards an optimal solution set of high-quality QA pairs.

Therefore, this methodology offers significant advantages over traditional dataset-creation methods. It provides automation and scalability, drastically reducing the time and cost associated with manual data generation while enabling the creation of large-scale, domain-specific datasets. The system also mitigates hallucinations through the use of Chain-of-Thought prompting and a rigorous evaluation and iterative refinement process, ensuring the factual accuracy and relevance of the generated question-answer pairs. Furthermore, the methodology ensures domain specificity by consistently using the source PDF document as context, tailoring the generated QA pairs to the information within. Diversity is promoted through the variation generation stage, exploring a broader range of potential questions and answers related to the source material. This framework represents a novel approach to leveraging the power of LLMs for automated, high-quality data generation. By integrating principles of genetic algorithms, the inherent limitations of LLMs are addressed, paving the way for the efficient and scalable creation of domain-specific datasets crucial for advancing the field of generative AI.
\section{Results and Discussion} \label{sect6}

\begin{figure*}[htp]
	\centering 
	\includegraphics[width=0.9\textwidth]{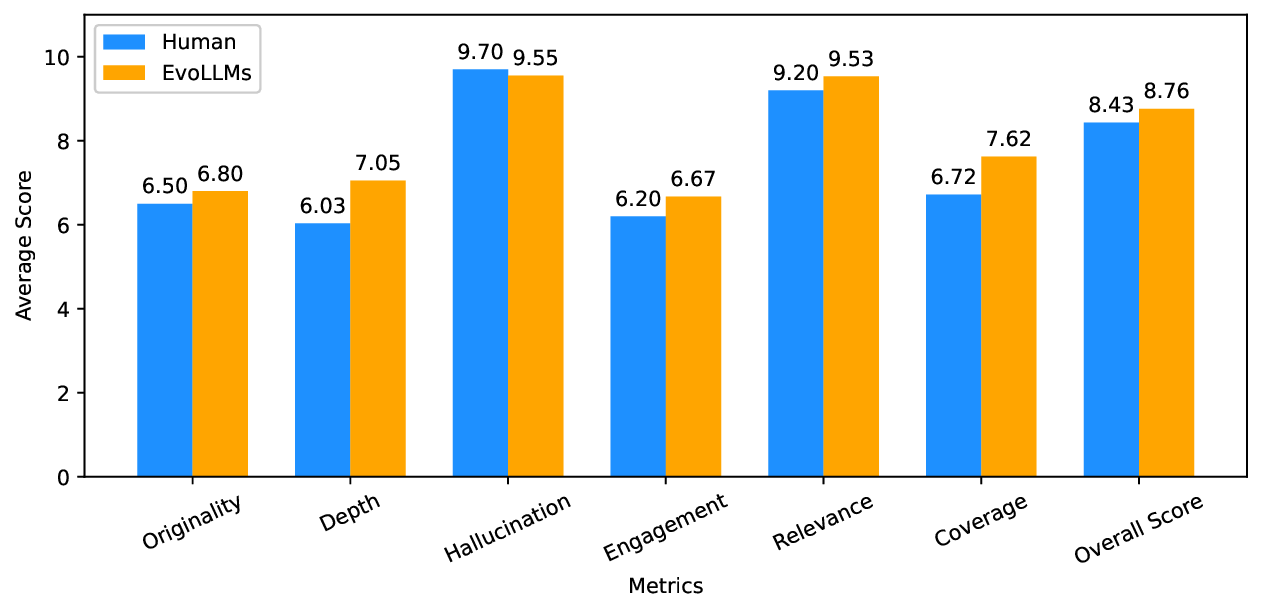} 
	\caption{Average Score per Metric (Human Expert vs. LLM).} 
	\label{fig6} 
\end{figure*}

The implementation of our proposed approach has been carried out on Google Colab\footnote{ \url{https://colab.google}}. We leverage free access to powerful GPUs, such as the NVIDIA T4, which greatly accelerates the training process. Additionally, we utilize a range of open-source platforms, including Hugging Face\footnote{\url{https://www.huggingface.co}}, LangChain\footnote{\url{https://www.langchain.com}}, and UNsloth\footnote{\url{https://www.unsloth.ai}}.

The results from the comparative analysis between the QA pairs generated by our EvoLLMs and those curated by human experts yielded significant insights, as visualized in Figure \ref{fig6}. Using a set of 15 distinct metrics, the evaluation process revealed that the EvoLLMs-generated dataset consistently outperformed the human expert-generated dataset in nearly all categories. The analysis demonstrates that EvoLLMs-generated dataset effectively mitigates hallucinations, nearly matching human performance in this category (9.55 vs. 9.70). Additionally, EvoLLMs outperforms human-generated datasets in several key areas, including Depth, Relevance, and Coverage, while maintaining a strong overall score (8.76 vs. 8.43). This balance indicates that EvoLLMs successfully reduces hallucinations without compromising the quality or relevance of the generated content, making it a robust solution for QA dataset generation.

Key metrics such as factual accuracy, depth of understanding, and clarity showed substantial improvement in the EvoLLMs-generated data. The chain-of-thought (CoT) prompting strategy proved to be particularly effective in guiding the model to generate coherent and logically structured responses, reducing the likelihood of hallucinations. This result aligns with our hypothesis that an evolutionary-inspired framework can enhance the overall quality of LLM outputs by incorporating iterative refinement and fitness evaluations.\\
In contrast, while accurate, the human expert-generated dataset exhibited limitations in diversity and adaptability. This was particularly evident in cases requiring nuanced phrasing and context interpretation variations, where the EvoLLMs showed greater flexibility. The variation generation stage of our methodology allowed the LLM to explore a broader range of possible question-answer pairs, enriching the dataset with diverse perspectives without compromising semantic integrity.\\
The computational efficiency of the EvoLLMs pipeline is another notable advantage. By automating the data generation process, we significantly reduced the time and resources typically required for manual curation. Using a feedback loop, where suboptimal QA pairs were iteratively refined, further contributed to the system's ability to converge on high-quality outputs with minimal human intervention.
\section{Conclusions and Future Work} \label{sect7}
In conclusion, this paper introduces a novel framework, EvoLLMs, that effectively addresses key challenges in QA dataset generation, particularly the mitigation of hallucinations. The proposed approach reduces hallucinations and enhances key metrics like depth, relevance, and coverage compared to human-generated datasets by leveraging evolutionary computation techniques, such as selection, variation, and mutation. The framework is computationally efficient and scalable, yielding diverse and accurate question-answer pairs. EvoLLMs' performance shows that the quality of generated outputs by LLMs may be drastically improved through automatized, iterative refinement procedures while decreasing dependence on painstakingly manual data curation.

While EvoLLMs has shown great potential in improving QA dataset generation and reducing hallucinations, a few directions remain for future work. First, the evolutionary algorithms can be further refined to make the system more adaptable across diverse domains, especially domains like healthcare and law, which are highly specialized and require factually accurate information. Integrating external knowledge sources, such as real-time databases or domain-specific ontologies, might improve the factual grounding of generated responses and reduce the risk of hallucinations.\\
The second area of future work would be expansion into the feedback mechanisms under EvoLLMs. The incorporation of more sophisticated evaluation metrics could include real-time user feedback or expert validation to help better align the model to the expectations of humans.

\bibliographystyle{IEEEtran}
\bibliography{IEEEabrv,refs}
\end{document}